\documentclass[11pt,a4paper]{article}
\usepackage[hyperref]{acl2020}
\usepackage{times}
\usepackage{latexsym}

\usepackage{microtype}
\usepackage{amssymb}
\usepackage{graphicx}
\usepackage{hyperref}
\usepackage{multirow}
\usepackage{adjustbox}
\usepackage{tabu}
\usepackage{amsmath}
\usepackage{algorithm,algpseudocode}
\usepackage{xcolor}
\usepackage{subcaption}

\aclfinalcopy

\title{Unsupervised Online Grounding of Natural Language during Human-Robot Interactions}

\author{Oliver Roesler \\
  Artificial Intelligence Lab \\
  Vrije Universiteit Brussel \\
  Brussels, Belgium \\
  \texttt{oliver@roesler.co.uk} \\}

\date{}

\begin{document}
\maketitle

\begin{abstract}
  Allowing humans to communicate through natural language with robots requires connections between words and percepts. The process of creating these connections is called symbol grounding and has been studied for nearly three decades. Although many studies have been conducted, not many considered grounding of synonyms and the employed algorithms either work only offline or in a supervised manner. In this paper, a cross-situational learning based grounding framework is proposed that allows grounding of words and phrases through corresponding percepts without human supervision and online, i.e. it does not require any explicit training phase, but instead updates the obtained mappings for every new encountered situation. The proposed framework is evaluated through an interaction experiment between a human tutor and a robot, and compared to an existing unsupervised grounding framework. The results show that the proposed framework is able to ground words through their corresponding percepts online and in an unsupervised manner, while outperforming the baseline framework. \\
\end{abstract}

\section{Introduction}\label{sec:introduction}
An increasing number of service robots is employed in human-centered complex environments and interacts with humans on a regular basis. This creates a need for robots that are able to understand instructions provided in natural language, such as \textit{bring a glas of water} or \textit{pick up a box}, to execute them appropriately and thereby enable efficient collaboration with humans. To this end, connections between words, i.e. abstract symbols, and their corresponding percepts, i.e. meanings, need to be created because according to the ''Symbol Grounding Problem'', which was proposed in 1990 by~\citet{harnad1990a}, abstract knowledge and language only has meaning, if it is linked to the physical world through mappings from words to corresponding percepts. \\
Grounding approaches can in general be separated into supervised and unsupervised approaches. The former utilize the guidance of a human tutor, while the latter do not require any supervision and try to use co-occurrence information to identify through which percepts a word is grounded. Previous studies, such as~\cite{kollar2010a,tellex2011a,aly2018a}, that investigated unsupervised grounding employed algorithms that only work offline, i.e. these algorithms need to be trained before deployment with in advance collected perceptual data and words, which prevents these algorithms from being used in real-time human-robot interactions. Additionally, most previous studies did not consider ambiguous words, although the sentences humans produce are often ambiguous due to homonymy, i.e. one word refers to several percepts, and synonymy, i.e. one percept can be referred to by several different words. The latter do not need to be true synonyms, i.e. words that refer to the exact same meaning, instead, words only need to be synonyms as references to a percept in a particular set of situations, e.g. \textit{coca cola} or \textit{lemonade} instead of \textit{bottle}. \\
In this paper, a recently proposed unsupervised online grounding framework~\cite{roesler2019c} is extended to handle real percepts obtained during human-robot interactions. More specifically, the learning framework is extended to first convert obtained percepts through clustering to an abstract representation, which is then used to ground all non-auxiliary words\footnote{Auxiliary words are words that do not have corresponding percepts and only exist for grammatical reasons.} of the encountered natural language instructions through cross-situational learning. Each shape, color, and action is referred to by at least two synonymous words, which need to be mapped to their corresponding geometric characteristics, color histograms, or kinematic features of the robot joints during action execution, to investigate the ability of the used frameworks to handle synonymous words. The grounding performance of the proposed framework is evaluated by comparing it to the grounding performance of a Bayesian grounding framework that has been used in several previous studies, e.g.~\cite{aly2018a,roesler2018a,roesler2019a}  \\
The rest of this paper is structured as follows: Sections (\ref{sec:background} and \ref{sec:relatedwork}) provide a brief overview of cross-situational learning and related work. Afterwards, an overview of the proposed unsupervised online grounding framework as well as the unsupervised Bayesian baseline framework is given in Sections (\ref{sec:groundingframework} and \ref{sec:baselineframework}). The experimental design and obtained results are described in Sections (\ref{sec:experimentalsetup} and~\ref{sec:resultsanddiscussion}). Finally, Section (\ref{sec:conclusionsandfuturework}) concludes the paper.

\section{Background}\label{sec:background}
Cross-situational learning (CSL) refers to the process of learning the meaning of words across multiple exposures to handle referential uncertainty. The basic idea is that a set of candidate meanings, i.e. mappings from words to percepts, can be created for every situation or context a word is used in and that the correct meaning can be obtained by determining the intersection of the sets of candidate meanings~\cite{pinker1989a,fisher1994a}. Thus, the correct mapping between a word and its corresponding percepts, i.e. its meaning, will reliably reoccur across situations~\cite{blythe2010a,smith2012b}.
A number of experimental studies have confirmed that humans use CSL for word learning, when no prior knowledge of language is available~\cite{akhtar1999a,gillette1999a,smith2008a}. Since CSL requires more than one exposure to learn a word, it belongs to the group of slow-mapping mechanisms through which most words are acquired~\cite{carey1978a}. In contrast, \textit{fast-mapping} allows words to be acquired through a single exposure, but it is only used for a limited number of words and can neither be explained nor achieved through CSL~\cite{carey1978b,vogt2012a}. Many different algorithms have been proposed to simulate CSL in humans and enable artificial agents, such as robots, to learn the meaning of words by grounding them through percepts (Section~\ref{sec:relatedwork}).

\section{Related Work}\label{sec:relatedwork}
Grounding is used to obtain the meaning of an abstract symbol, e.g. a word, by linking it to perceptual information, i.e. the ``real'' world~\cite{harnad1990a}. There exist many different approaches for grounding. \citet{she2014a} grounded higher level symbols through already grounded lower level symbols with the help of a dialog system. Since the system requires a sufficiently large set of grounded lower level symbols as well as  a professional tutor to answer its questions, its usefulness is limited. The need for a human tutor that knows the correct mappings also limits the applicability of the \textit{Naming Game}, which allows an agent to quickly learn word-percept mappings~\cite{steels2012b}. In contrast to the previous approaches, cross-situational learning (Section~\ref{sec:background}), which assumes that one word appears several times together with the same perceptual feature vector so that a corresponding mapping can be created, does not require a human tutor for grounding~\cite{siskind1996a,smith2011a}. Previous studies investigated the use of cross-situational learning for grounding of objects, actions, and spatial concepts~\cite{roesler2019a,dawson2013a}. In most studies, grounding was conducted offline, i.e. perceptual data and words were collected in advance, which prevents these approaches from being used in real-time human-robot interactions. In contrast to these approaches, the framework used in this study learns the correct mappings from words to percepts online while interacting with humans and does not require separate training and test phases. Additionally, the majority of employed models were not able to handle ambiguous words, although, the sentences humans produce are often ambiguous due to homonymy and synonymy. One recent study showed that grounding of known synonyms does not require semantic or syntactic information and that such information can even have a negative effect, depending on the characteristics of the used information and how it is applied~\cite{roesler2018a}. Therefore, the online grounding mechanism employed in this study uses no additional semantic or syntactic information to ground synonyms.
 
\section{Grounding Framework}\label{sec:groundingframework}
The employed grounding framework consists of four parts: (1) 3D object segmentation component, which segments objects into point clouds to determine their geometric characteristics and colors, (2) Action recording component, which creates action feature vectors by recording the states of several joints while the robot is executing actions, (3) Percept clustering component, which obtains an abstract representation of percepts through clustering, and (4) Cross-situational learning based grounding component, which identifies auxiliary words and maps percepts to non-auxiliary words and phrases. The inputs and outputs of the individual parts are highlighted below, and described in detail in the following subsections.

\begin{enumerate}
  \item \textbf{3D object segmentation}:
    \begin{itemize}
      \item \textbf{Input}: Point cloud data.
      \item \textbf{Output}: Geometric characteristics and colors of objects.
    \end{itemize}
  \item \textbf{Action recording}:
    \begin{itemize}
      \item \textbf{Input}: Changes of the robot's joint states during action execution.
      \item \textbf{Output}: Action feature vectors representing the executed actions.
    \end{itemize}
  \item \textbf{Clustering of percepts}:
    \begin{itemize}
      \item \textbf{Input}: Geometric object characteristics, object colors, and action feature vectors.
      \item \textbf{Output}: Cluster numbers of percepts.
    \end{itemize}
  \item \textbf{Cross-situational learning}:
    \begin{itemize}
      \item \textbf{Input}: Natural language instructions and cluster numbers of percepts.
      \item \textbf{Output}: Word to percept mappings.
    \end{itemize}
\end{enumerate}

\subsection{3D Object Features}\label{subsec:3dobjectfeatures}
\begin{figure}[tb]
  \centering
  \includegraphics[width=0.48\textwidth,clip,keepaspectratio]{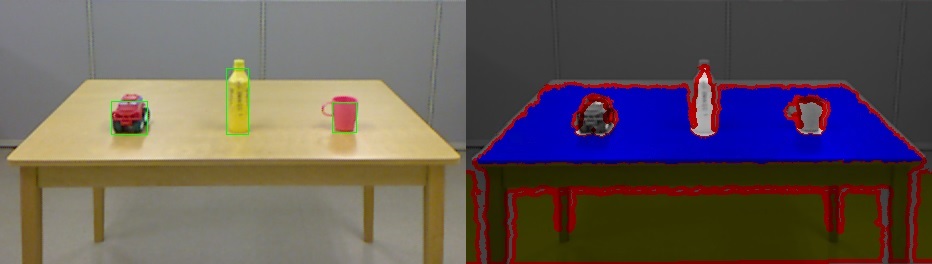}
  \caption{Illustration of the used objects and the corresponding 3D point cloud information: (A) car, (B) bottle, and (C) cup.}
  \label{fig:objectsoverview}
\end{figure}

In this study, an unsupervised model based 3D point cloud segmentation approach is used to segment objects lying in a plane into separate point clouds because it is fast, reliable and does not need much prior knowledge, such as object models or the number of regions to process~\cite{craye2016a}. The applied model uses the RANSAC algorithm~\cite{fischler1981a} to detect the major plane in the environment, which is a tabletop in the conducted experiment, and keeps track of it in consecutive frames. If a plane is orthogonal to the major plane and touches at least one border of the image, it is defined as a wall plane. After filtering out points that belong to the main plane or wall planes, the remaining points are voxelized and clustered into blobs representing object candidates. Blobs that are neither extremely small nor large are treated as objects\footnote{The threshold for the blob size was manually set based on the objects used in the experiment and should be suitable for all objects of similar size.}. Point clouds of segmented objects are characterized through Viewpoint Feature Histogram (VFH) descriptors~\cite{rusu2010a}, which represent the object geometries taking into consideration the viewpoints while ignoring scale variances, and color histograms, which represent the colors of the objects. Figure (\ref{fig:objectsoverview}) provides an illustrative example of the obtained 3D point cloud information.

\subsection{Action Features}\label{subsec:actionfeatures}
Action feature vectors are used to represent the dynamic characteristics of actions during execution through teleoperation. Overall, five different characteristics, which represent possible subactions, are recorded through the sensors of the robot~\cite{toyotamotorcorporation2017a}. The used characteristics are:

\begin{enumerate}
  \item The distance from the actual to the lowest torso position in meters.
  \item The angle of the arm flex joint in radians.
  \item The angle of the wrist roll joint in radians.
  \item Velocity of the base.
  \item Binary state of the gripper (1: closing, 0: opening or no change).
\end{enumerate}

They are then combined into the following vector:
\[
  \begin{pmatrix}
    a_{1}^{1} & a_{1}^{2} & a_{1}^{3} & a_{1}^{4} & a_{1}^{5} \\
    \vdots & \vdots & \vdots & \vdots & \vdots \\
    a_{6}^{1} & a_{6}^{2} & a_{6}^{3} & a_{6}^{4} & a_{6}^{5}
  \end{pmatrix},
\]
where \(a^{1}\) represents the difference of the distances from the lowest torso position in meters, while \(a^{2}\) and \(a^{3}\) represent the differences in the angles of the arm flex and wrist roll joints in radians, respectively. The differences are calculated by subtracting the values at the beginning of the subaction from the values at the end of the subaction. \(a^{4}\) represents the mean velocity of the base (forward/backward), and \(a^{5}\) represents the binary gripper state. Each action is characterized through six manually defined subactions. Therefore, if an action consists of less than six subactions, rows with zeros are added at the end, while the duration of a subaction is not fixed because it depends on the teleoperator.

\subsection{Clustering of percepts}\label{subsec:perceptsclustering}
The CSL algorithm (Section~\ref{subsec:crosssituationallearning}) requires percepts to be converted to an abstract representation that can then be used to ground natural language. The abstract representation is obtained through clustering as proposed in~\cite{roesler2019e}. Since it cannot be assumed that the number of clusters, i.e. the number of different percepts, is known in advance, DBSCAN, which is a density-based clustering algorithm proposed by~\citet{ester1996a}, is used\footnote{The used DBSCAN implementation is available in \textit{scikit-learn}~\cite{pedregosa2011a}.} because it determines the number of clusters automatically, while only requiring two parameters, i.e. the radius \(\epsilon\) and threshold \textit{minSamples}. Each iteration DBSCAN determines a number of core points, which are points that have more than \textit{minSamples} points within radius \(\epsilon\) around them~\cite{schubert2017a}. All the points within radius \(\epsilon\) of a core point are assigned to the same cluster as the core point. \\ Cluster numbers are calculated every situation prior to grounding so that they can be provided to the CSL algorithm. Recalculating them every situation is necessary to take into account the new percepts of that situation.

\begin{algorithm}[tb]
  \caption{The grounding procedure takes as input all words (\textit{W}) and percepts (\textit{P}) of the current situation, the sets of all previously obtained word-percept (\textit{WP}) and percept-word (\textit{PW}) pairs, the set of auxiliary words (AW), and the set of permanent phrases (PP) and returns the sets of grounded words (\textit{GW}) and percepts (\textit{GP}).}
  \begin{algorithmic}[1]
    \Procedure{Grounding}{\textit{W}, \textit{P}, \textit{WP}, \textit{PW}, \textit{AW}, \textit{PP}}
    \State Substitute words with phrases from \textit{PP}
    \State Update \textit{AW} (Algorithm~\ref{algo:awdetection}) and remove \textit{AW} from \textit{W}
    \State Update \textit{WP} and \textit{PW} using \textit{W} and \textit{P}
    \For {$j = 1$ to $word\_number$}
      \State Save highest \textit{WP} to \textit{GW}
    \EndFor
    \For {$j = 1$ to $percept\_number$}
      \State Save highest \textit{PW} to \textit{GP}
      \EndFor
    \State \textbf{return} $GW \cup GP$
    \EndProcedure
  \end{algorithmic}
  \label{algo:grounding}
\end{algorithm}

\begin{algorithm}[t]
  \caption{The auxiliary word detection procedure takes as input the sets of word and percept occurrences (\textit{WO} and \textit{PO}),  and the set of detected auxiliary words (AW).}
  \begin{algorithmic}[1]
    \Procedure{Auxiliary Word Detection}{\textit{WO}, \textit{PO}, \textit{AW}}
    \For {word, occurrence in \textit{WO}}
      \If {$occurrence > max(PO)*2$}
        \State Add word to \textit{AW}
      \EndIf
    \EndFor
    \State \textbf{return} \textit{AW}
    \EndProcedure
  \end{algorithmic}
  \label{algo:awdetection}
\end{algorithm}

\subsection{Cross-Situational Learning}\label{subsec:crosssituationallearning}
A variety of algorithms have been developed that realize CSL in different ways, e.g. through the use of probabilistic models~\cite{aly2018a,roesler2019a}, to ground words through percepts in artificial agents. This section describes an online CSL algorithm for grounding of words, which has first been proposed by~\citet{roesler2018b} and recently been extended with auxiliary word and phrase detection~\cite{roesler2019c}. Since the sentences in this study are shorter, have a much simpler structure, and less variation than the sentences used in~\cite{roesler2019c}, the previous auxiliary word and phrase detection algorithms do not work. Thus, a novel auxiliary word detection algorithm (Algorithm~\ref{algo:awdetection}) is proposed to handle the simpler sentences employed in this study\footnote{Both auxiliary word mechanisms, i.e. the one used in~\cite{roesler2019c} and the one proposed in this study, are used in parallel because both have shown to not produce false detections, i.e. they either detect an auxiliary word correctly or do not detect it.}, while no phrase detection is used to ensure a fair comparison with the baseline framework (Section~\ref{sec:baselineframework}), which does not have any phrase detection capabilities. The rest of this section provides an overview of the employed grounding algorithm. \\
For each situation all corresponding words and percepts are given to the grounding algorithm (Algorithm~\ref{algo:grounding}), while the sets of grounded words (\textit{GW}) and percepts (\textit{GP}) are initially empty. Before the actual grounding procedure, words that are part of known phrases will be combined so that they can be grounded together and auxiliary words are automatically detected and removed (Algorithm~\ref{algo:awdetection}). Afterwards, all possible word-percept (\textit{WP}) and percept-word (\textit{PW}) pairs are created, i.e. for each word and percept a set containing all percepts and words they occurred with is created, and saved together with a number indicating how often the pair occurred. The highest word-percept pair is determined and saved to the set of grounded words (\textit{GW}). All other word-percept pairs the word or percept are part of will no longer be considered for the selection of the highest word-percept pair in future iterations. This restriction is applied until all percepts have been used once for grounding. Afterwards, if some words have not been grounded, all percepts will become again available for grounding until all words have been grounded to allow grounding of synonyms. After all words have been grounded the same process is repeated for percept-word pairs to assign synonymous percepts to the same word. Finally, the sets of grounded words and percepts are merged. \\

\section{Baseline Framework}\label{sec:baselineframework}
The baseline framework consists of three parts: (1) 3D object segmentation component as described in Section (\ref{subsec:3dobjectfeatures}), (2) Action recording component as described in Section (\ref{subsec:actionfeatures}), and (3) Bayesian learning model, which identifies auxiliary words and grounds non-auxiliary words and phrases through corresponding percepts. Since the perceptual data extraction components are the same for both frameworks, any difference in grounding performance can only be due to the different grounding algorithms, i.e. component three and four of the proposed framework (Sections~\ref{subsec:perceptsclustering} and~\ref{subsec:crosssituationallearning}) and component three of the baseline framework, which is described in the remainder of this section.

\begin{figure}[tb]
  \centering
  \includegraphics[width=0.49\textwidth,clip,keepaspectratio]{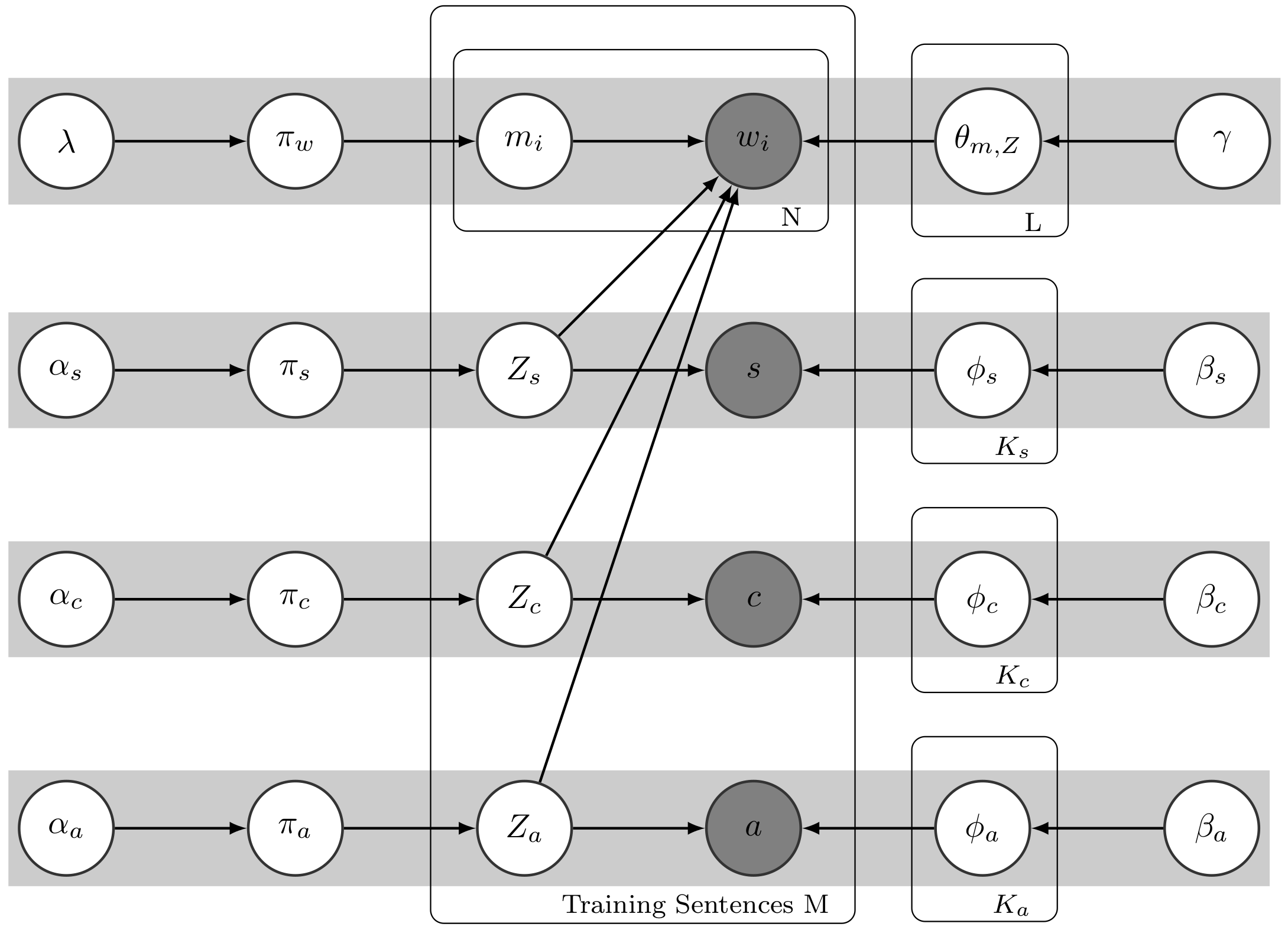}
  \caption{Graphical representation of the probabilistic model. Indices \textbf{i}, \textbf{s}, \textbf{c}, and \textbf{a} denote the order of words, object shapes, object colors, and actions, respectively.}
  \label{fig:baselineframework-graphicalmodel}
\end{figure}

\begin{table}[tb]
  \centering
  \caption{Definitions of the learning parameters in the graphical model.}
  \adjustbox{max width=0.49\textwidth}{
    \begin{tabu}{|[1.5pt gray]c|[1.5pt gray]c|[1.5pt gray]}
      \hline
      \textbf{Parameter} & \textbf{Definition} \\ \hline
      \(\lambda\) & Hyperparameter of the distribution \(\pi_{w}\) \\ \hline
      \(\alpha_{s}, \alpha_{c}, \alpha_{a}\) & Hyperparameters of the distributions \(\pi_{s}, \pi_{c}\) and \(\pi_{a}\) \\ \hline
      \multirow{2}{*}{\(m_{i}\)} & Modality index of each word \\
        & (modality index \(\in\) \{Shape, Color, Action, AW\}) \\ \hline
      \(Z_{s}, Z_{c}, Z_{a}\) & Indices of shape, color and action distributions \\ \hline
      \(w_{i}\) & Word indices \\ \hline
      \(s, c, a\) & Observed states representing shapes, colors and actions \\ \hline
      \(\gamma\) & Hyperparameter of the distribution \(\theta_{m,Z}\) \\ \hline
      \(\beta_{s}, \beta_{c}, \beta_{a}\) & Hyperparameters of the distributions \(\phi_{s}, \phi_{c}\) and \(\phi_{a}\) \\ \hline
      \(\theta_{m,Z}\) & Word distribution over modalities \\ \hline
    \end{tabu}
  }
  \label{tab:baselineframework-parameterdefinitions}
\end{table}

The probabilistic learning model, described in this section, is based on the model used in~\cite{roesler2019a}, since the experimental setup employed in this study (Section~\ref{sec:experimentalsetup}) is also based on the scenario used in~\cite{roesler2019a}. In general, the model has been chosen as a baseline because similar models have previously been employed in similar grounding scenarios by different researchers, e.g.~\cite{kollar2010a,tellex2011a,aly2018a,roesler2018a,roesler2019a}. In the model, the observed state \(w_{i}\) represents word indices, i.e. each individual word is represented by a different integer\footnote{The following two example sentences illustrate the representation of words through word indices: (please, \textbf{1}) (lift up, \textbf{2}) (the, \textbf{3}) (brown, \textbf{4}) (coke, \textbf{5}) and (lift up, \textbf{2}) (the, \textbf{3}) (brownish, \textbf{6}) (lemonade, \textbf{7}), where the bold numbers indicate word indices.}. The observed state \textit{s} represents the shape of objects, more specifically their geometric characteristics expressed through VFH descriptors (Section~\ref{subsec:3dobjectfeatures}), \textit{c} represents the color of objects and \textit{a} represents actions. Table (\ref{tab:baselineframework-parameterdefinitions}) provides a summary of the definitions of the learning model parameters. The corresponding probability distributions, i.e., \(w_{i}\), \(\theta_{m,Z_{L_{1}}}\), \(\phi_{s_{K_{1}}}\), \(\phi_{c_{K_{2}}}\), \(\phi_{a_{K_{3}}}\), \(\pi_{w}\), \(\pi_{s}\), \(\pi_{c}\), \(\pi_{a}\), \(m_{i}\), \(Z_{s}\), \(Z_{c}\), \(Z_{a}\), \textit{s}, \textit{c}, and \textit{a}, which characterize the different modalities in the graphical model, are defined in Equation (\ref{eqn:baselineframework-probabilitydistributions}), where \textit{Cat} denotes a categorical distribution, \textit{Dir} denotes a Dirichlet distribution, \textit{GIW} denotes a Gaussian Inverse-Wishart distribution, and \textit{N} denotes a multivariate Gaussian distribution.

\begin{equation}
  \left\{
    \begin{array}{lllll}
      w_{i} & \sim & Cat(\theta_{m_{i}, Z_{m_{i}}}) &  \\
      \theta_{m,Z_{L_{1}}} & \sim & Dir(\gamma)\hspace{0.5cm} , & L_{1} = (1,...,L) \\
      \phi_{s_{K_{1}}} & \sim & GIW(\beta_{s})\hspace{0.1cm}, & K_{1} = (1,...,K_{s}) \\
      \phi_{c_{K_{2}}} & \sim & GIW(\beta_{c})\hspace{0.1cm}, & K_{2} = (1,...,K_{c}) \\
      \phi_{a_{K_{3}}} & \sim & GIW(\beta_{a})\hspace{0.1cm}, & K_{3} = (1,...,K_{a}) \\
      \pi_{w} & \sim & Dir(\lambda) &  \\
      \pi_{s} & \sim & Dir(\alpha_{s}) &  \\
      \pi_{c} & \sim & Dir(\alpha_{c}) &  \\
      \pi_{a} & \sim & Dir(\alpha_{a}) &  \\
      m_{i} & \sim & Cat(\pi_{w}) &  \\
      Z_{s} & \sim & Cat(\pi_{s}) &  \\
      Z_{c} & \sim & Cat(\pi_{c}) &  \\
      Z_{a} & \sim & Cat(\pi_{a}) &  \\
      s & \sim & N(\phi_{Z_{s}}) &  \\
      c & \sim & N(\phi_{Z_{c}}) &  \\
      a & \sim & N(\phi_{Z_{a}}) &  \\
    \end{array}
  \right.
  \label{eqn:baselineframework-probabilitydistributions}
\end{equation}

The latent variables of the Bayesian learning model are inferred using the Gibbs sampling algorithm~\cite{geman1984a} (Algorithm~\ref{algo:baselineframework-gibbssampling}), which repeatedly samples from and updates the posterior distributions (Equation~\ref{eqn:baselineframework-posteriordistributions}). Distributions were sampled for 100 iterations, after which convergence had been achieved.

\begin{equation}
  \left\{
    \begin{array}{lll}
      \phi_{s} & \sim & P(\phi_{s}|s, \beta_{s}) \\
      \phi_{c} & \sim & P(\phi_{c}|c, \beta_{c}) \\
      \phi_{a} & \sim & P(\phi_{a}|a, \beta_{a}) \\
      \pi_{w} & \sim & P(\pi_{w}|\lambda, m) \\
      \pi_{s} & \sim & P(\pi_{s}|\alpha_{s}, Z_{s}) \\
      \pi_{c} & \sim & P(\pi_{c}|\alpha_{c}, Z_{c}) \\
      \pi_{a} & \sim & P(\pi_{a}|\alpha_{a}, Z_{a}) \\
      Z_{s} & \sim & P(Z_{s}|s, \pi_{s}, w) \\
      Z_{c} & \sim & P(Z_{c}|c, \pi_{c}, w) \\
      Z_{a} & \sim & P(Z_{a}|a, \pi_{a}, w) \\
      \theta_{m, Z} & \sim & P(\theta_{m,Z}|m, Z_{s}, Z_{c}, Z_{a}, \gamma, w) \\
      m_{i} & \sim & P(m_{i}|\theta_{m, Z}, Z_{s}, Z_{c}, Z_{a}, \pi_{w}, w_{i}) \\
    \end{array}
  \right.
  \label{eqn:baselineframework-posteriordistributions}
\end{equation}

\section{Experimental Setup}\label{sec:experimentalsetup}
The experimental scenario used in this study is based on the scenario used in~\cite{roesler2019a}. The main difference is the use of an additional modality, i.e. color, which leads to slightly different sentences.  During the experiment a human tutor and HSR robot\footnote{The Human Support Robot from Toyota, which is used for the experiment, has an omnidirectional movable cylindrical shaped body with one arm and gripper. It is equipped with a variety of different sensors, such as stereo and wide-angle cameras, and has 11 degrees of freedom.} interact in front of a table, with one of the following five objects $\boldsymbol\{$\textsc{bottle}, \textsc{cup}, \textsc{box}, \textsc{car}, and \textsc{book}$\boldsymbol\}$ (Figure~\ref{fig:objectsoverview}). Each interaction follows below procedure: \\

\begin{enumerate}
  \item The human tutor places an object on the table and the robot determines the object's geometric characteristics and color to create corresponding feature vectors (Section~\ref{subsec:3dobjectfeatures}).
  \item An instruction, which describes how to manipulate the object, is given to the robot by the human tutor, e.g. ``please lift up the red soda''.
  \item The human tutor teleoperates the robot to execute the action provided through the instruction while several kinematic characteristics are recorded and converted into an action feature vector (Section~\ref{subsec:actionfeatures}).
\end{enumerate}

\begin{algorithm}[tb]
  \caption{Inference of the model's latent variables. In this study, \(nr\_of\_iterations\) was set to 100.}
  \begin{algorithmic}[1]
    \Procedure{Gibbs sampling}{\textit{W}, \textit{P}, \textit{WP}, \textit{AW}}
    \State Initialization of \(\theta, \phi_{s}, \phi_{c}, \phi_{a}, \pi_{w}, \pi_{s}, \pi_{c},\)
    \State \(\pi_{a}, Z_{s}, Z_{c}, Z_{a}, m_{i}\)
    \For {\(i = 1\) to \(nr\_of\_iterations\)}
      \State Equation ($\ref{eqn:baselineframework-posteriordistributions}$)
      \EndFor
      \State \textbf{return} \(\theta, \phi_{s}, \phi_{c}, \phi_{a}, \pi_{w}, \pi_{s}, \pi_{c}, \pi_{a}, Z_{s},\)
      \State \phantom{return} \(Z_{c}, Z_{a}, m_{i}\)
    \EndProcedure
  \end{algorithmic}
  \label{algo:baselineframework-gibbssampling}
\end{algorithm}

\begin{figure}[tb]
  \centering
  \includegraphics[width=0.49\textwidth,clip,keepaspectratio]{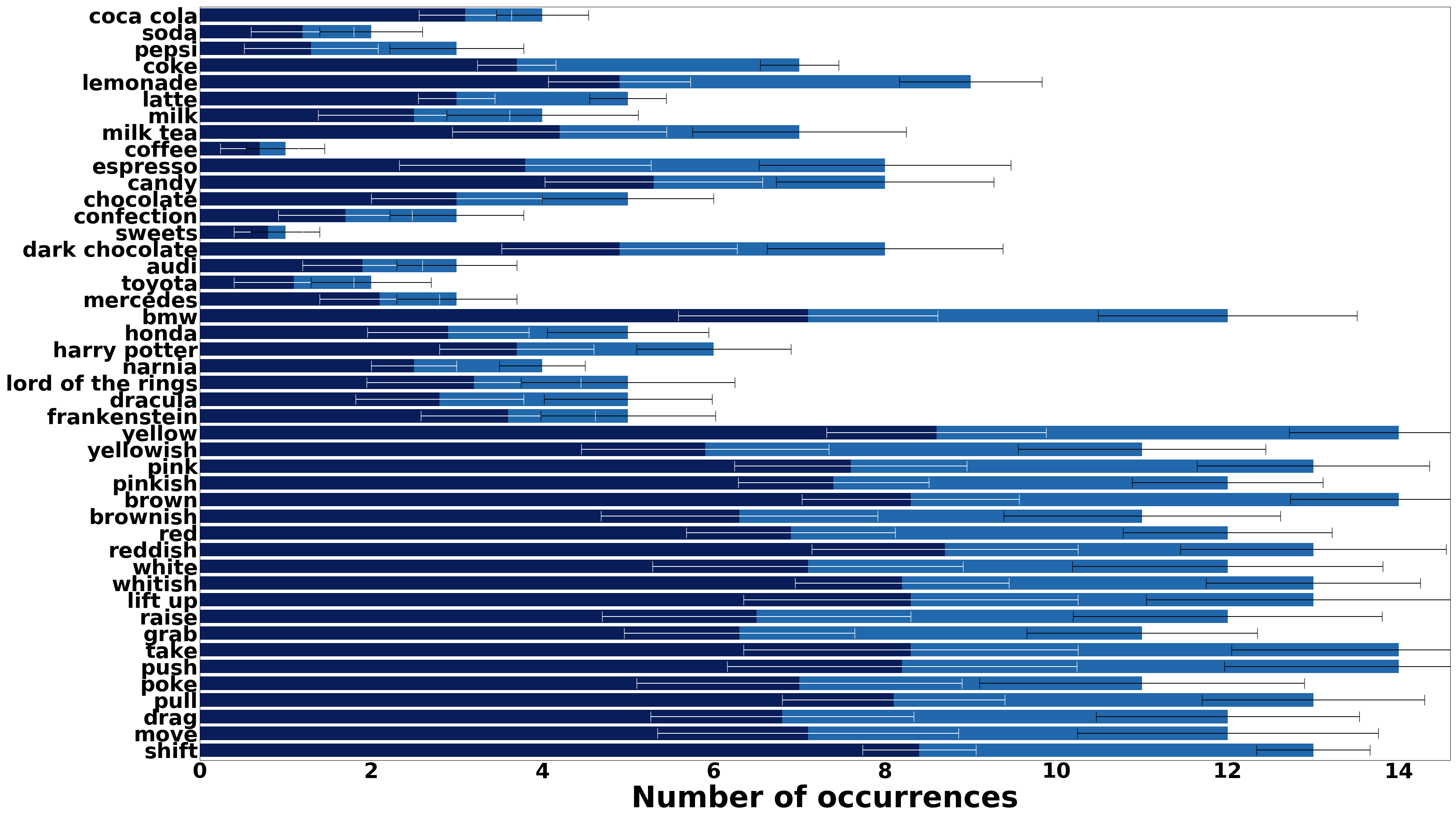}
  \caption{Word occurrences for all encountered words except auxiliary words. The dark blue part of the bars shows the mean number of occurrences during training and the bright blue part the mean number of occurrences during testing.}
  \label{fig:wordoccurrences}
\end{figure}

A total of 125 interactions were performed to record perceptual information for all combinations of employed shapes, colors, and actions. Since instruction words were selected randomly for each situation, except that words had to fit the encountered percepts, their number of occurrences in the data varies, e.g. the word ``coffee'' only occurs once, while the word ``brown'' occurs 14 times. Grounding was then performed for ten different interaction sequences, i.e. the order of the recorded situations was randomly changed, to ensure that the performance is not due to the specific order in which situations are encountered. Figure (\ref{fig:wordoccurrences}) shows how often each word occurred on average in all interactions as well as the training and test interactions. \\
Each sentence consists of one of the following structures: ``\textit{action} the \textit{color} \textit{shape}'' or ``please \textit{action} the \textit{color} \textit{shape}'', where \textit{action}, \textit{color}, and \textit{shape} are substituted by one of their corresponding words (Table~\ref{tab:perceptsandwordsoverview}). Each action and color can be referred to by two different words, while each shape has five corresponding words. During training and testing the obtained situations are given to the proposed and baseline frameworks. The former framework gets the situations separately one after the other, as if it is processing the data in real-time during the interaction. It first clusters the percepts of the current situation together with all previously encountered percepts to obtain abstract representations of shapes, colors and actions (Section~\ref{subsec:perceptsclustering}). Afterwards, the CSL based grounding algorithm is used to ground words through the obtained cluster numbers (Section~\ref{subsec:crosssituationallearning}). In contrast, the baseline framework does not allow online learning and requires all sentences and corresponding percepts of the training situations to be given at once to the learning model.

\begin{table}[tb]
  \centering
  \caption{Overview of all percepts with their corresponding synonyms. The action percepts are explained in Table (\ref{tab:actionsoverview}).}
  \adjustbox{max width=0.48\textwidth}{
    \begin{tabu}{|[1.5pt gray]c|[1.5pt gray]c|[1.5pt gray]c|[1.5pt gray]c|[1.5pt gray]c|[1.5pt gray]c|[1.5pt gray]}
      \hline
      \bf{Type} & \bf{Percept} & \bf{Synonyms} \\ \hline
      \multirow{5}{*}{Shape} & Bottle & coca cola, soda, pepsi, coke, lemonade \\ \cline{2-3}
        & Cup & latte, milk, milk tea, coffee, espresso \\ \cline{2-3}
        & Box & candy, chocolate, confection, sweets, dark chocolate \\ \cline{2-3}
        & Car & audi, toyota, mercedes, bmw, honda \\ \cline{2-3}
        & Book & harry potter, narnia, lord of the rings, dracula, frankenstein \\ \hline
      \multirow{5}{*}{Color} & Yellow & yellow, yellowish \\ \cline{2-3}
        & Pink & pink, pinkish \\ \cline{2-3}
        & Brown & brown, brownish \\ \cline{2-3}
        & Red & red, reddish \\ \cline{2-3}
        & White & white, whitish \\ \hline
      \multirow{5}{*}{Action} & Lift up & lift up, raise \\ \cline{2-3}
        & Grab & grab, take \\ \cline{2-3}
        & Push & push, poke \\ \cline{2-3}
        & Pull & pull, drag \\ \cline{2-3}
        & Move & move, shift \\ \hline
      \multirow{2}{*}{\shortstack{Auxiliary\\ Word}} & - & the \\ \cline{2-3}
        & - & please \\ \hline
    \end{tabu}
  }
  \label{tab:perceptsandwordsoverview}
\end{table}

\begin{table}[tb]
  \centering
  \caption{Explanations of the employed action percepts.}
  \adjustbox{max width=0.48\textwidth}{
    \begin{tabu}{|[1.5pt gray]c|[1.5pt gray]c|[1.5pt gray]}
      \hline
      \bf{Percept} & \bf{Description} \\ \hline
      Lift up & The object will be grabbed and lifted up. \\ \hline
      Grab & The object will be grabbed, but not displaced. \\ \hline
      Push & The object will be pushed with the closed gripper without being grabbed first. \\ \hline
      Pull & The object will be grabbed and moved towards the robot. \\ \hline
      Move & The object will be grabbed and moved away from the robot. \\ \hline
    \end{tabu}
  }
  \label{tab:actionsoverview}
\end{table}

\begin{figure}[tb]
  \centering
  \includegraphics[width=0.49\textwidth,clip,keepaspectratio]{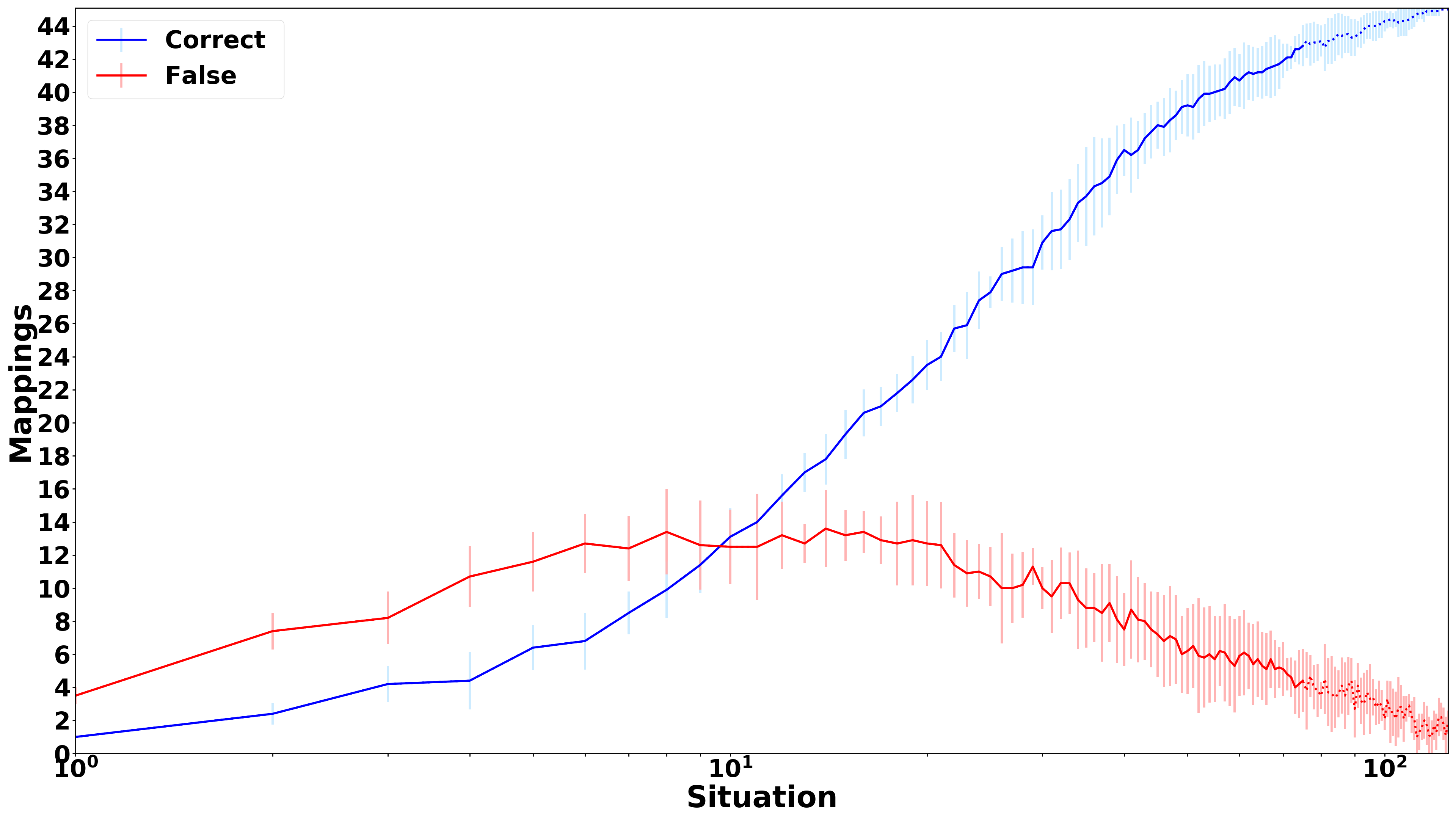}
  \caption{Mean number and standard deviation of correct and false mappings obtained by the proposed model over all 125 situations. The dotted part only occurs, when all situations are used for training, otherwise the model obtains only 43 correct mappings.}
  \label{fig:mappings}
\end{figure}

\begin{figure}[tb]
  \centering
  \begin{subfigure}[h]{0.49\textwidth}
    \centering
    \includegraphics[width=\textwidth,clip,keepaspectratio]{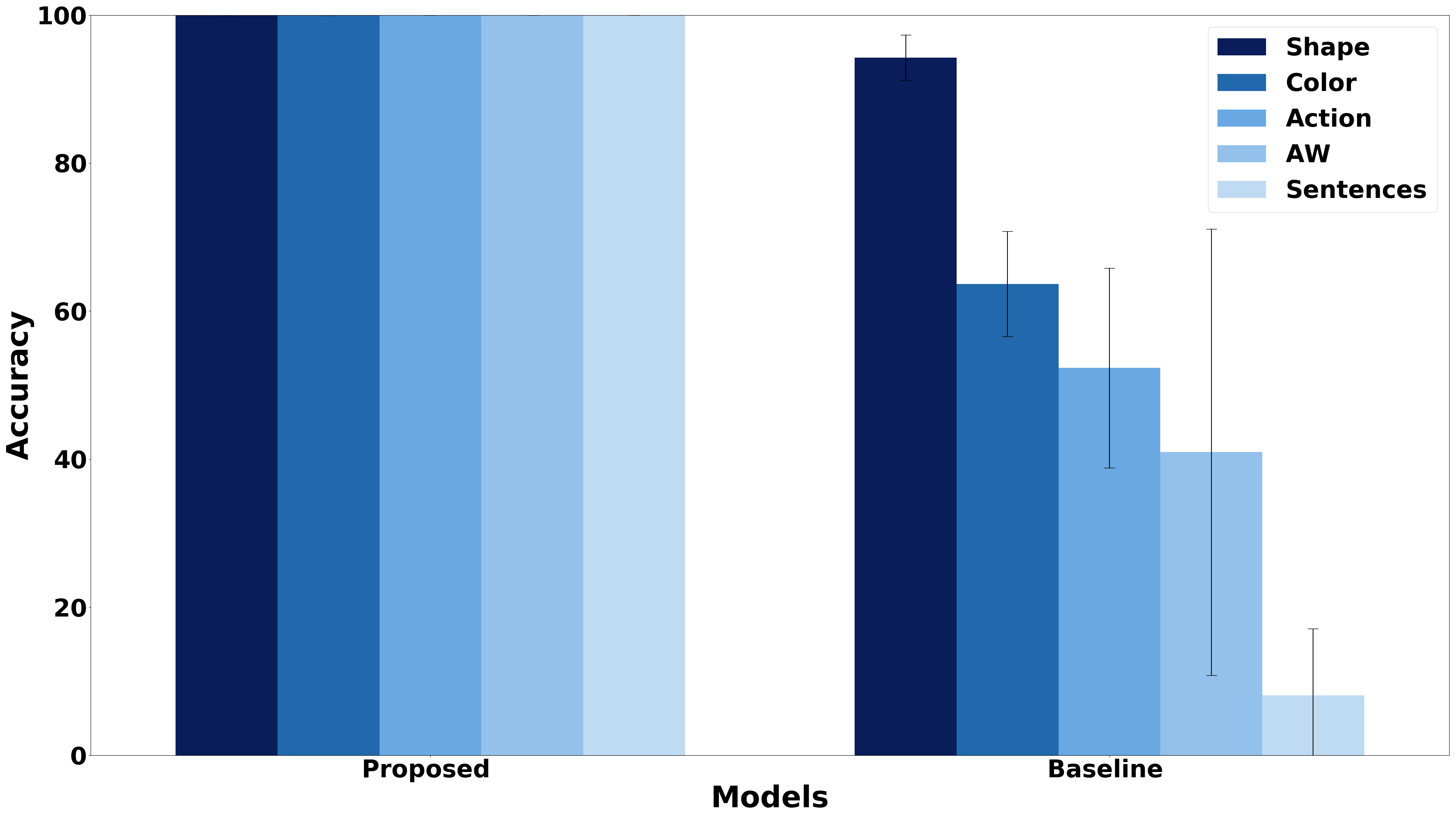}
    \caption{All situations used for training and testing.}
    \label{subfig:accuracymodalitiessentences1.0}
  \end{subfigure}
  \begin{subfigure}[h]{0.49\textwidth}
    \centering
    \includegraphics[width=\textwidth,clip,keepaspectratio]{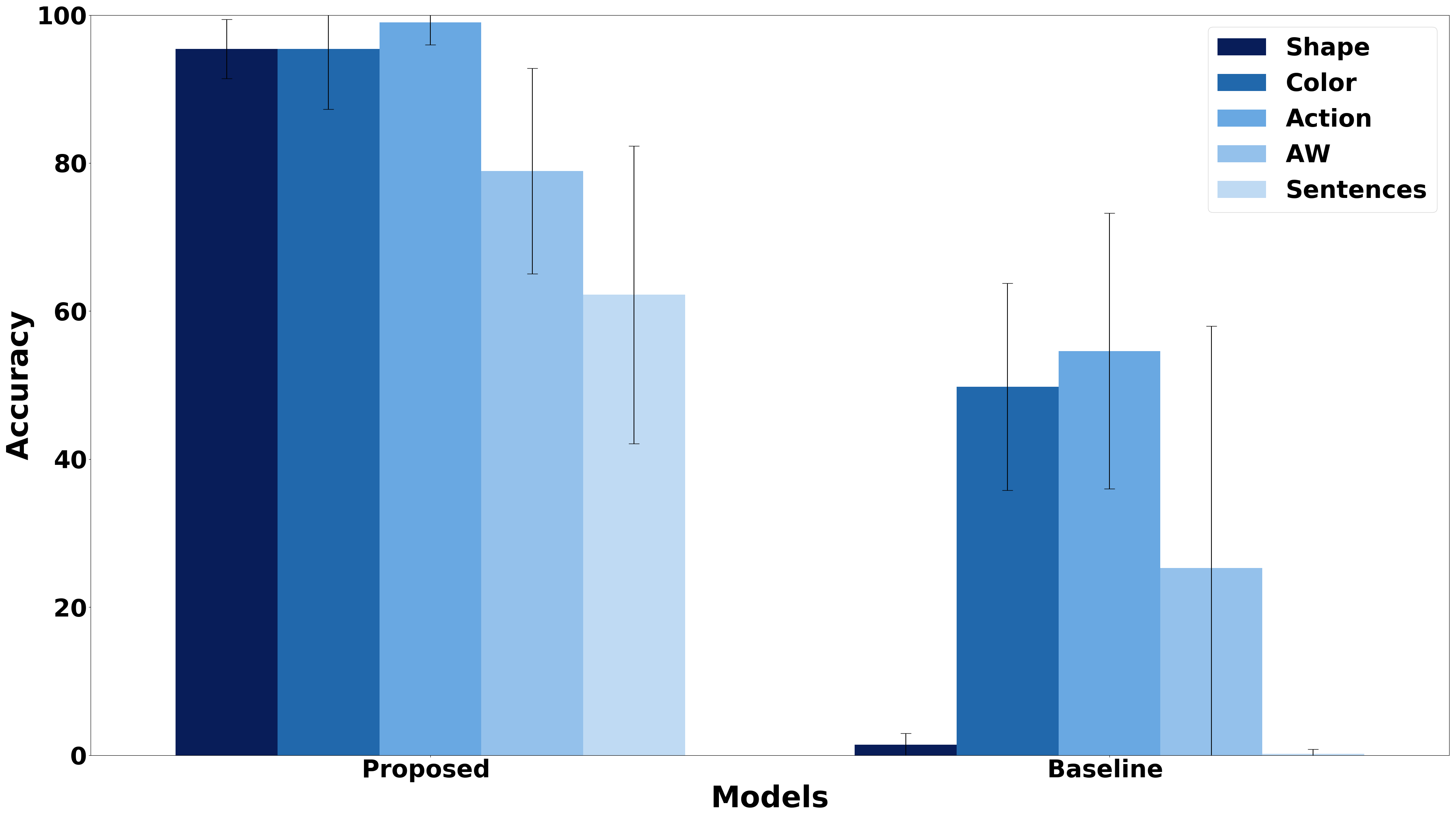}
    \caption{60\% of the situations used for training and 40\% for testing.}
    \label{subfig:accuracymodalitiessentences0.6}
  \end{subfigure}
  \caption{Mean grounding accuracy results and corresponding standard deviations for all modalities and both models. Additionally, the percentage of sentences for which all words were correctly grounded is shown.}
  \label{fig:accuracymodalitiessentences}
\end{figure}

\section{Results and Discussion}\label{sec:resultsanddiscussion}
The proposed cross-situational learning framework (Section~\ref{sec:groundingframework}) is evaluated through a human-robot interaction scenario (Section~\ref{sec:experimentalsetup}) and the obtained grounding results are compared to the groundings achieved by an unsupervised Bayesian grounding framework (Section~\ref{sec:baselineframework}). Figure (\ref{fig:mappings}) shows how the mean number of correct and false mappings changes, when the proposed grounding framework encounters the employed situations one after the other. It also shows that all 45 correct mappings are obtained, when all 125 situations are used for training, while on average only 43 correct mappings are obtained, when only 60\% of the situations are used for training. The figure also illustrates the online grounding capability of the model, i.e. that it updates its mappings with every new encountered situation, as well as its transparency because it allows to check at any time through which percept a word is grounded at that moment. Based on the collected co-occurrence information it would also be possible to calculate a confidence score for every mapping to understand how likely it is that a false mapping disappears or a correct mapping persists. The described transparency of the proposed framework can be helpful to understand and debug responses to instructions provided by a human, when the framework is used to control an artifical agent interacting with a human, especially when the responses are incorrect or inappropriate.

\begin{figure*}[tb]
  \centering
  \begin{subfigure}[h]{0.49\textwidth}
    \centering
    \includegraphics[width=\textwidth,clip,keepaspectratio]{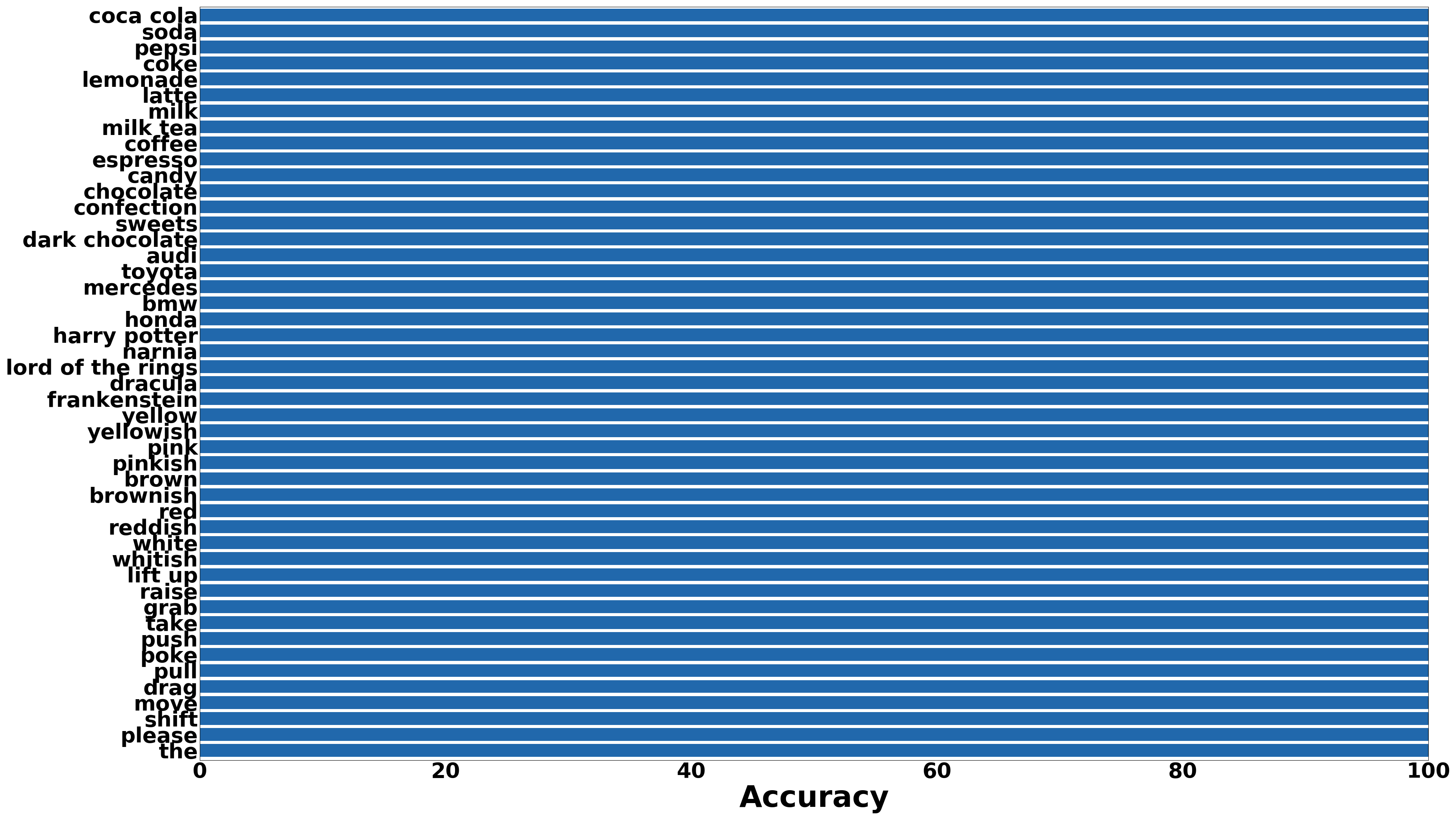}
    \caption{Proposed model using all situations for training and testing.}
    \vspace{0.2cm}
    \label{subfig:accuracywordsproposed1.0}
  \end{subfigure}
  \begin{subfigure}[h]{0.49\textwidth}
    \centering
    \includegraphics[width=\textwidth,clip,keepaspectratio]{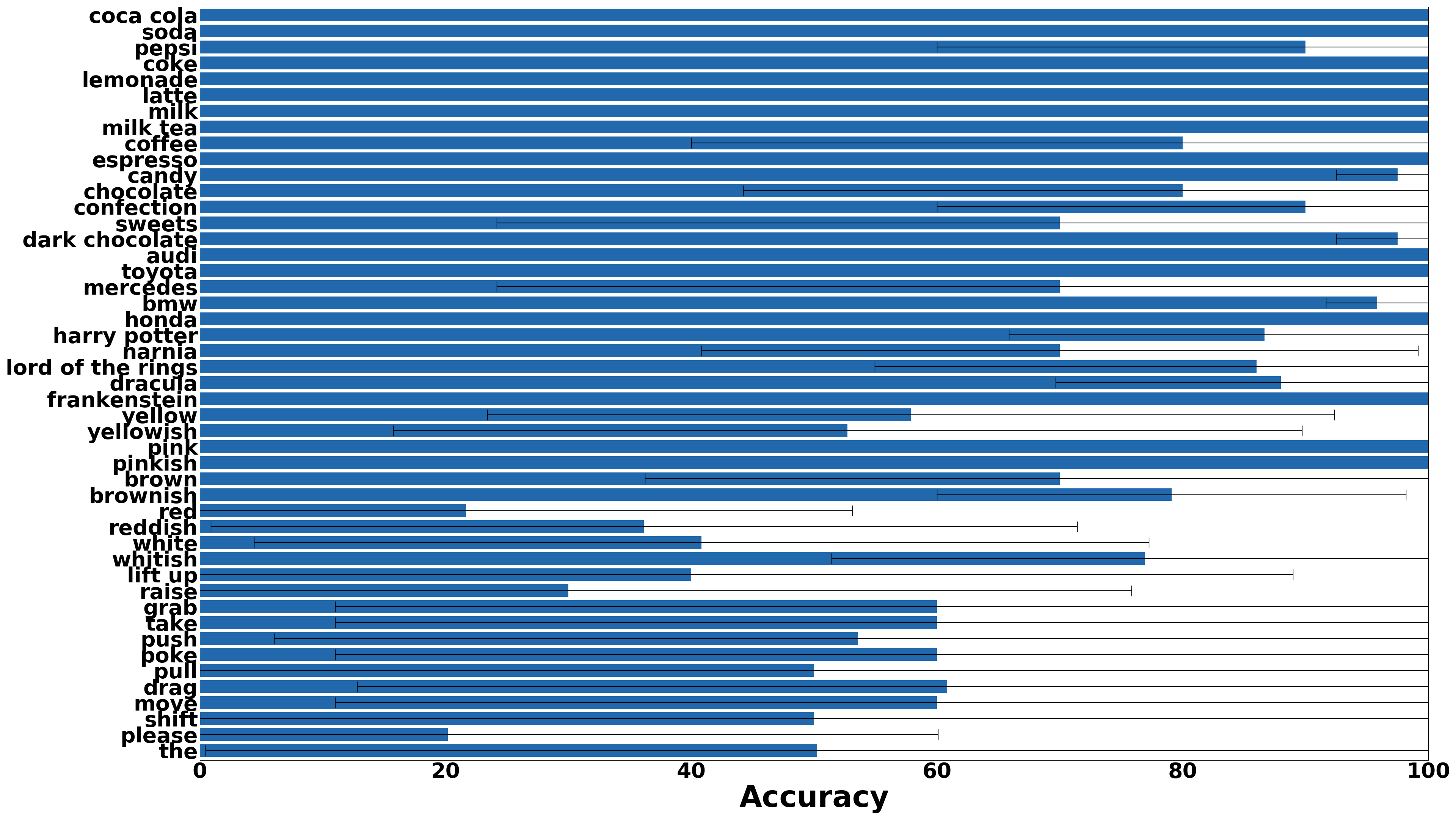}
    \caption{Baseline model using all situations for training and testing.}
    \vspace{0.2cm}
    \label{subfig:accuracywordsbaseline1.0}
  \end{subfigure}
  \begin{subfigure}[h]{0.49\textwidth}
    \centering
    \includegraphics[width=\textwidth,clip,keepaspectratio]{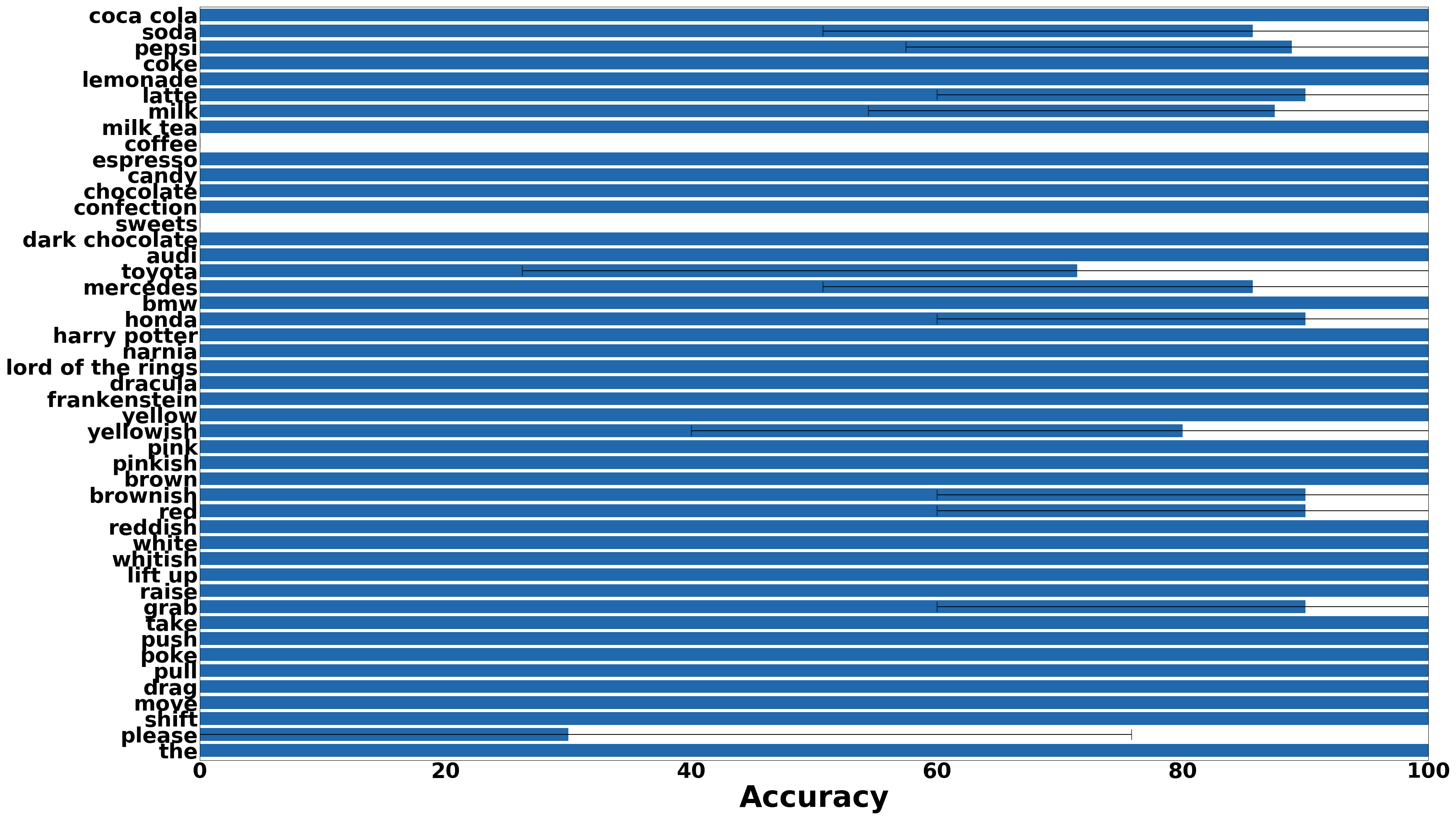}
    \caption{Proposed model using 60\% of the situations for training and 40\% for testing.}
    \label{subfig:accuracywordsproposed0.6}
  \end{subfigure}
  \begin{subfigure}[h]{0.49\textwidth}
    \centering
    \includegraphics[width=\textwidth,clip,keepaspectratio]{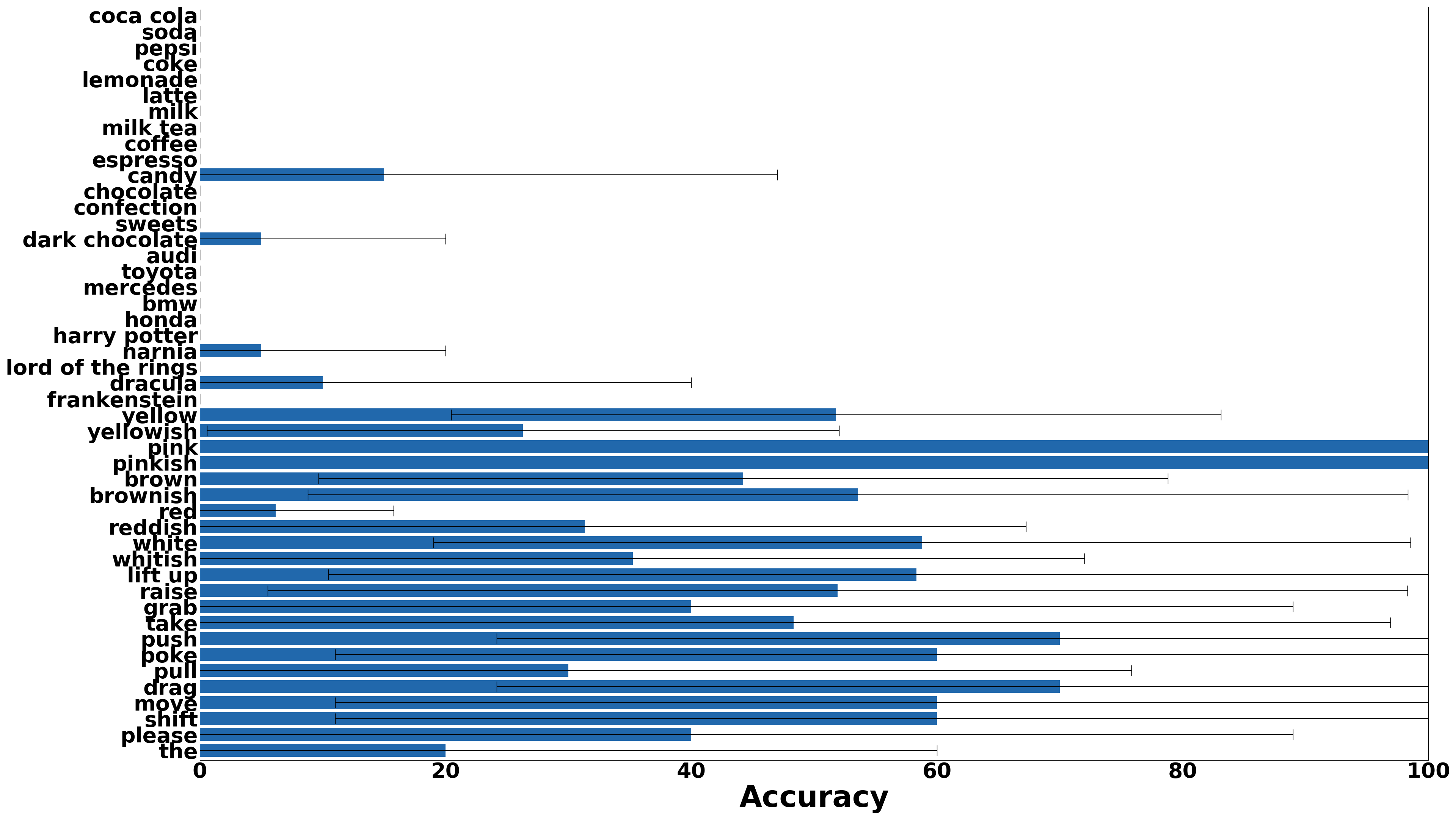}
    \caption{Baseline model using 60\% of the situations for training and 40\% for testing.}
    \label{subfig:accuracywordsbaseline0.6}
  \end{subfigure}
  \caption{Mean accuracy results and corresponding standard deviations for each individual word.}
  \label{fig:accuracywords}
\end{figure*}

In contrast, the baseline model requires an explicit training phase so that no corresponding figure, illustrating the number of correct and false mappings, can be created. Thus, to allow a comparison between the two models, the mappings of the proposed model are extracted after 125 and 75 situations, depending on the used train/test split. Two different train/test splits are analyzed in this study. For the first split, all situations are used for training and testing to see how well the frameworks perform when all test situations have been encountered before. For the second split, only 60\% of the used situations are provided for training, while the remaining situations are used for testing. In this case, it is possible that some words never occur during training or only a limited number of times, e.g. once or twice. If a word does not occur during training, the proposed model is not able to obtain a corresponding mapping which leads to an accuracy of 0\% as shown in Figure (\ref{subfig:accuracywordsproposed0.6}) for the words \textit{coffee} and \textit{sweets}, which both only exist once in the dataset and are thus only present during training or testing, but not both. The word accuracies shown in Figure (\ref{fig:accuracywords}) were calculated by dividing the number of times a word was correctly grounded through the number of times the word was encountered during testing. Similar to the proposed model, the baseline model was also not able to ground the words \textit{coffee} and \textit{sweets} correctly, when only 60\% of the situations were used for training. However, the baseline model also seems to require in general a higher minimum number of occurrences to successfully ground words, since there are many words that achieved a mean accuracy of 0\%, when only 60\% of the situations were used for training (Figure~\ref{subfig:accuracywordsbaseline0.6}). \\
Figure (\ref{subfig:accuracymodalitiessentences1.0}) shows that the proposed model achieves perfect grounding, when the same situations are provided for training and testing, which confirms that it is able to obtain all correct mappings as shown in Figure (\ref{fig:mappings}). However, if only 60\% of the situations are used for training and the remaining 40\% unknown situations for testing the grounding accuracy drops for both models. For the proposed model the largest accuracy decrease is seen for auxiliary words, while still more than 95\% of the obtained shape, color and action groundings are correct. For the baseline framework the largest drop in accuracy is seen for shapes, from more than 95\% to less than 2\%. The reason might be that every shape word has 5 synonyms, thus, if words would be equally distributed among all situations and specifically among the training and test sets, the decrease might not be as sharp. However, Figure~\ref{fig:wordoccurrences} shows that the number of occurrences is not necessarily the reason for the drop because the words \textit{bmw} and \textit{narnia} occured on average 7 and 2.5 times during training, respectively, and \textit{narnia} achieved an accuracy of about 5\%, while the accuracy of \textit{bmw} was 0\% (Figure~\ref{subfig:accuracywordsbaseline0.6}). In contrast, the proposed model shows a more stable performance, since it was able to ground all non-auxiliary words that occured at least one time during training with a mean accuracy of more than 70\%, while only the auxiliary word \textit{please} achieved a lower mean accuracy of 30\%. \\
Overall the evaluation shows that the proposed model outperforms the baseline model based on its auxiliary word detection and grounding accuracy. Interestingly, the performance difference is larger, when only 60\% of the situations are used for training, although this scenario is artificially harming the proposed model by preventing it to learn during testing, since it does not require explicit training. In addition to the better grounding performance, the proposed model is also more transparent, which becomes important when robots are interacting with humans in complex and unrestricted environments, especially if some actions of the robots can cause harm to humans.

\section{Conclusions and Future Work}\label{sec:conclusionsandfuturework}
This paper investigated a multimodal framework for grounding synonymous shape, color and action words through the visual perception and proprioception of a robot during its interaction with a human tutor. The cross-situational learning model was set up to learn the meaning of shape and color words of objects as well as action words using geometric characteristics and color information of objects obtained from point cloud information as well as kinematic features of the robot joints recorded during action execution. \\
The proposed model allowed auxiliary word detection and online grounding of synonyms through real percepts in an unsupervised manner and without the use of any syntactic or semantic information. Additionally, it outperformed the baseline model based on the accuracy of the obtained groundings, its capability to process new situations online and its transparency. \\
In future work, different mechanisms will be investigated to improve the sample efficiency of the algorithm, which will become relevant, if a larger number of words is used or words occur less often. Additionally, it will be verified whether the framework can handle homonyms. Finally, supervised grounding methods will be integrated so that the robot is able to use human feedback, but does not require it.

\bibliography{paper}

\end{document}